\documentclass[10pt,conference]{IEEEtran}
\usepackage[utf8]{inputenc}
\usepackage{graphicx}
\usepackage{amsmath}
\usepackage{hyperref}

\usepackage{algorithm}
\usepackage{algpseudocode}
\algrenewcommand\algorithmicrequire{\textbf{Input:}}
\algrenewcommand\algorithmicensure{\textbf{Output:}}

\usepackage{tikz}
\usetikzlibrary{arrows.meta,positioning,fit,calc,shapes.misc}
\usepackage{graphicx,booktabs,makecell,array}
\newcolumntype{C}[1]{>{\centering\arraybackslash}m{#1}} % centered fixed-width column

\title{%
Mitigating Memorization in Text-to-Image Diffusion via Region-Aware Prompt Augmentation and Multimodal Copy Detection

}
\author{
Yunzhuo Chen$^{1}$, Jordan Vice$^{1}$, Naveed Akhtar$^{2,1}$, Nur Al Hasan Haldar$^{3,1}$, Ajmal Mian$^{1}$ \\
$^{1}$The University of Western Australia, Perth, Australia \\
$^{2}$The University of Melbourne, Melbourne, Australia \\
$^{3}$Curtin University, Perth, Australia \\
{\tt\small yunzhuo.chen@research.uwa.edu.au, naveed.akhtar1@unimelb.edu.au,} \\
{\tt\small nur.haldar@curtin.edu.au, jordan.vice@uwa.edu.au, ajmal.mian@uwa.edu.au}
}

\begin{document}
\maketitle

\begin{abstract}
State-of-the-art text-to-image diffusion models can produce impressive visuals but may memorize and reproduce training images, creating copyright and privacy risks. Existing prompt perturbations applied at inference time, such as random token insertion or embedding noise, may lower copying but often harm image-prompt alignment and overall fidelity. To address this, we introduce two complementary methods. First, Region-Aware Prompt Augmentation (RAPTA) uses an object detector to find salient regions and turn them into semantically grounded prompt variants, which are randomly sampled during training to increase diversity, while maintaining semantic alignment. Second, Attention-Driven Multimodal Copy Detection (ADMCD) aggregates local patch, global semantic, and texture cues with a lightweight transformer to produce a fused representation, and applies simple thresholded decision rules to detect copying without training with large annotated datasets. Experiments show that RAPTA reduces overfitting while maintaining high synthesis quality, and that ADMCD reliably detects copying, outperforming single-modal metrics.
\end{abstract}

\section{Introduction}
Diffusion models have rapidly become the leading approach for high‐fidelity image synthesis. Despite their achievements, diffusion models can copy their training data, a behavior documented across small- and large-scale retrieval works \cite{kim2023mitigating, somepalli2022diffusion}.  Such replication raises legal and ethical concerns, from inadvertent copyright infringement to privacy leaks. Additionally, the subtle “reconstructive memory” of semantically equivalent objects complicates the boundary between creative synthesis and forgery. We interpret copying as arising from the combination of large model capacity, strong text–image alignment, and over‐reliance on training‐time caption–image pairs.  

In practice, large overparameterized models are trained on web-scale, weakly curated pairs of captions and images. Strong text–image alignment encourages the model to anchor on specific caption–image pairs, and high capacity enables memorization of distinctive instances. Together, these factors can surface both near-duplicate reproduction and style-level echoes of training items \cite{kim2023mitigating,somepalli2023memorize}. Although replication is often deemed undesirable, it manifests along a spectrum ranging from exact pixel-level duplication to stylistic or semantic mimicry, and in certain contexts (e.g., data augmentation or style transfer), controlled copying may be beneficial.  This motivates a more nuanced view in which replication is graded rather than binary.

\begin{figure}[t]
\centering
\includegraphics[width=0.9\linewidth]{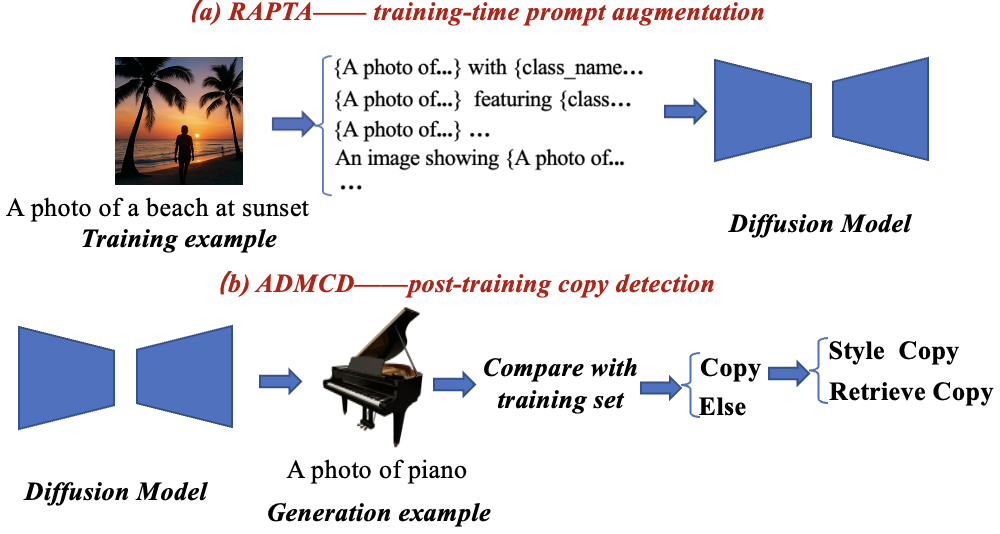}
\caption{\textbf{Overview of our framework.}
\textbf{(a) RAPTA} augments the training caption in an image-grounded way: a detector provides region proposals with classes and coarse positions; a small template set instantiates region-aware variants from the base prompt. One variant is sampled per iteration to condition the conditional diffusion model, injecting grounded diversity while preserving prompt–image alignment.
\textbf{(b) ADMCD} is an inference-time, two-step detector. For a generated image $G$, we compare it against the training set using fused-feature cosine similarity $S_{\mathrm{fus}}$; pairs with $S_{\mathrm{fus}}>\tau_1$ are flagged as Copy. A similarity-weighted score over three streams (visual, CLIP, texture) with threshold $\tau_2$ assigns the copy type.}
\label{fig:intro}
\end{figure}

Existing mitigation works often modifies prompts at inference time. Examples include random token or number insertion, BLIP-based paraphrasing \cite{li2022blip}, and adding Gaussian noise to CLIP embeddings \cite{kim2023mitigating}. These  approaches can reduce copying, but they weaken prompt–image coherence and do not address memorization during training. Regarding detection, single-view measures like SSIM, copy-move forensics (SSCD) \cite{pizzi2022sscd}, or CLIP cosine similarity \cite{radford2021learning} provide only coarse signals and depend on class diversity and human judgment. Progress is also limited by the lack of large, labeled copy pair datasets tailored to diffusion models.

To address these limits, we introduce two complementary modules: Region-Aware Prompt Augmentation (RAPTA) and Attention-Driven Multimodal Copy Detection (ADMCD). RAPTA diversifies prompts during training with object-centric templates. It employs a pretrained detector \cite{ren2015fasterrcnn} on each training image to obtain high-confidence regions, encoding object class and coarse spatial layout details into a set of prompt variants - randomly sampling one variant per iteration. This exposes the model to semantically grounded descriptions without damaging semantic alignment. ADMCD extracts three feature streams: (i) patch-level embeddings from a Vision Transformer \cite{dosovitskiy2021vit}, (ii) global text-conditioned embeddings from CLIP \cite{radford2021clip} and, (iii) texture descriptors from a ResNet backbone \cite{he2016resnet}. These features are projected into a shared latent space and fused with a lightweight transformer. Exploiting cosine-similarity \cite{manning2008ir} thresholds on the fused vector enable zero-shot copy detection. Together, RAPTA and ADMCD provide an end-to-end pipeline for mitigating and detecting unintended memorization in text-to-image diffusion models.

\begin{figure*}[t]
\centering
\includegraphics[width=1\linewidth]{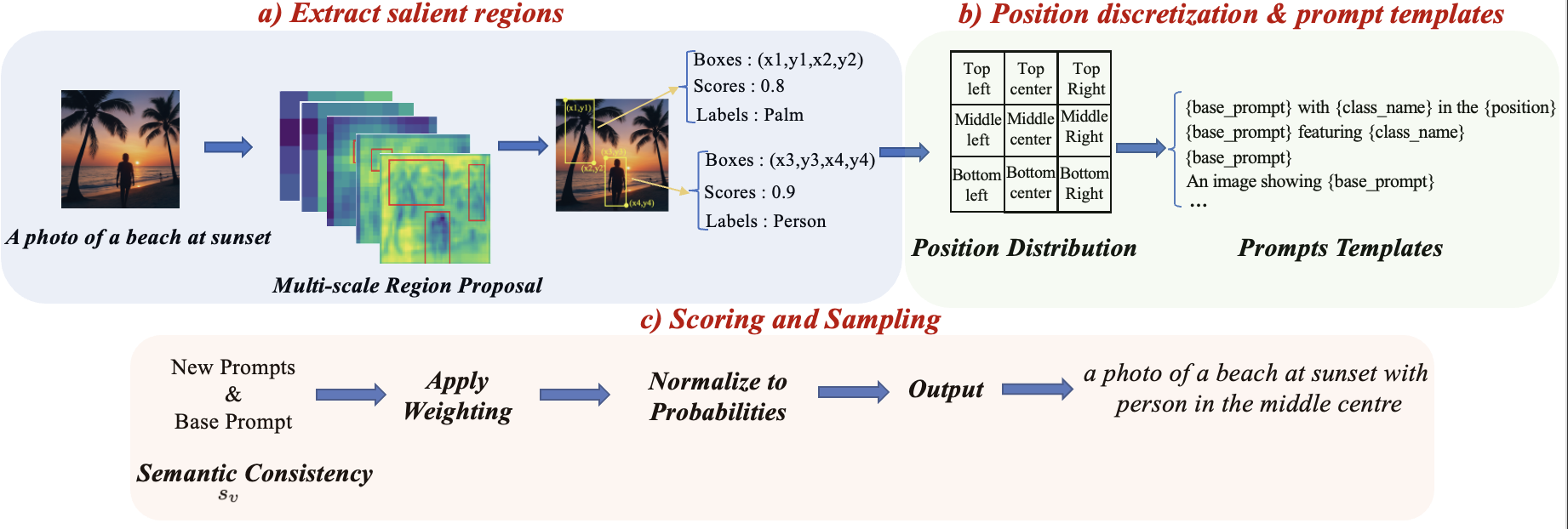}
\caption{\textbf{RAPTA pipeline.}
\textbf{(a)} A pretrained detector highlights saliency-based boxes with class/confidence; we keep the top-$M$ with $S_i>\tau_b$ and discretize each box center to a $3\times3$ grid $\mathcal{G}$ (top-left $\cdots$ bottom-right), yielding position tokens $\mathrm{pos}_i$.
\textbf{(b)} From the base prompt $p$, detected class $c_i$ and position $\mathrm{pos}_i$, a small template set $\{T_j\}_{j=1}^{J}$ instantiates region-aware variants $V=\{p\}\cup\{\,T_j(p,c_i,\mathrm{pos}_i)\,\}$.
\textbf{(c)} For image $I$ and each $v\in V$, CLIP consistency $S_v=\cos\!\big(f_I,f_v\big)$ is converted to weights $w_v=S_v^{\gamma}$ and normalized to $\pi(v)=w_v/\sum_{u\in V}w_u$; we sample a single prompt $\tilde p\sim\pi(\cdot)$ to condition training. This preserves prompt–image alignment while injecting grounded diversity.}
\label{fig:rapta_pipeline}
\end{figure*}

We evaluate our approach on LAION‐10k using several pre‐trained diffusion backbones and detection baselines.  While our measured copy rates on LAION‐10k likely understate real‐world prevalence, RAPTA consistently enhances prompt–image alignment under diverse training, and ADMCD delivers robust, zero‐training copy detection.  Our contributions are:
\begin{itemize}
  \item Region-Aware Prompt Augmentation (RAPTA): a training-time augmentation scheme that turns detector proposals into object-aware prompt variants. It discretizes region centers to a coarse grid, instantiates a small template set, and uses a CLIP-based sampling policy with temperature to select one variant per iteration. This yields semantically grounded diversity with negligible overhead and without degrading text–image alignment.;
  \item Attention-Driven Multimodal Copy Detection (ADMCD): a lightweight detector that aggregates three streams—local patch, global semantic, and texture. A two-stage rule is used: a copy decision threshold on the fused cosine similarity, and a copy-type threshold on a weighted stream score to separate retrieve/exact from style copies. The detector operates without task-specific training data.
  \item ADMCD as a similarity measure: beyond detection, the fused similarity provides a more accurate notion of image similarity than most current methods, aligning better with human judgments and remaining stable under common photometric and geometric attacks, as shown by quantitative and qualitative studies.
  \item Experiments on three conditional diffusion backbones and a 1.2k-pair evaluation set show lower copy rates with RAPTA at comparable or better FID/KID, and superior detection and robustness with ADMCD. Ablations cover threshold calibration, weight selection, and corruption robustness.
\end{itemize}

\section{Related Work}

Denoising diffusion probabilistic models (DDPM) introduced the idea of learning to reverse a gradual corruption process via deep neural networks, achieving high‐fidelity image synthesis \cite{ho2020denoising}.  Latent diffusion further improves efficiency by operating in a compressed latent space, enabling high‐resolution generation on modest hardware \cite{rombach2022high}.  Text‐conditional pipelines such as Stable Diffusion extend these advances to natural language prompts, producing diverse and photorealistic outputs from text. Recent studies show that text–to–image diffusion models can reproduce training images or their distinctive attributes, at both small and web scale \cite{somepalli2022forgery,somepalli2023memorize,kim2023mitigating,carlini2023extracting}.This has raised concerns among researchers about copyright issues.

Most practical defenses act at inference time by altering the \emph{prompt text} or its embedding: random token/number insertion, BLIP-based paraphrasing \cite{li2022blip}, or Gaussian noise on CLIP embeddings \cite{kim2023mitigating}. These methods can lower copying but often weaken prompt–image coherence and do not address training-time memorization. Training-time ideas include caption diversification or object/layout conditioning (e.g., GLIGEN/ControlNet-style grounding \cite{yuan2024object,zhang2023controlnet}), but generic templates or paraphrases can introduce semantic drift. RAPTA differs by making diversification \emph{region-aware}: detector proposals and coarse grid positions instantiate a small, semantically grounded template set that is CLIP-scored and randomly sampled per iteration.

Digital-image forensics for duplication spans three families.  First, classical copy--move methods detect duplicated regions via local features and block-based analysis (e.g., SIFT/SURF/ORB) and are often summarized by an inlier ratio after geometric verification \cite{rublee2011orb}.  Second, learned copy-detection descriptors such as SSCD yield a global, transformation-robust fingerprint; detection is performed by nearest-neighbor retrieval in the training set followed by cosine-similarity thresholding \cite{pizzi2022sscd}. Third, a large body of work adapts perceptual similarity metrics as detectors by applying a unified “metric-as-detector’’ protocol: compute a score between a query and candidate references and make a binary decision via a threshold. Common choices include SSIM for structural similarity \cite{wang2004ssim}, LPIPS for deep-feature perceptual distance \cite{zhang2018lpips}, DISTS that unifies structure and texture \cite{ding2021dists}, DreamSim that better captures human-perceived similarity \cite{fu2023dreamsim}, and CLIP-based global-semantic cosine used widely for retrieval-style detection \cite{radford2021learning}. 

While effective and scalable, these single-view scores struggle with partial or stylistic copying, are sensitive to photometric or geometric perturbations, and cannot reliably distinguish retrieve/exact copies from style-level mimicry; thresholds also vary across categories and domains. Motivated by these limitations, our ADMCD keeps the standard scoring-and-threshold decision rule but replaces the score with an attention-fused, multimodal similarity that jointly aggregates patch-level geometry, global semantics, and texture cues, and further introduces a second threshold to separate retrieve/exact from style copying.

Image similarity metrics can be brittle under photometric or geometric changes. Robustness benchmarks such as ImageNet-C \cite{hendrycks2019imagenetc} highlight sensitivity to noise, blur, and weather-like artifacts. Our experiments show that the fused similarity in ADMCD remains stable across ten common image-based attacks (noise/blur/Poisson/salt–pepper/speckle, crop/flip/occlude/rotate), whereas single-view metrics vary widely, supporting ADMCD as a stronger similarity measure in this setting.

\section{Method}

Collecting large, reliably labeled sets of generated–reference image pairs for training robust classifiers is prohibitively labor‐intensive, and human visual inspection remains the ultimate arbiter of copying.  Moreover, single‐metric detection cannot distinguish nuanced forms of content reuse. To address these gaps, we propose two complementary modules: Region‐Aware Prompt Augmentation (RAPTA)
% , a training‐time prompt diversification method grounded in detected object regions that preserves prompt–image alignment while increasing diversity;
and Attention‐Driven Multimodal Copy Detection (ADMCD), a lightweight transformer‐based detector multi-level features for robust copy detection.
% that fuses patch‐level, global semantic, and texture features to deliver robust copy detection without requiring large labeled corpora.

\subsection{Region-Aware Prompt Augmentation (RAPTA)}
\label{sec:rapta}

As illustrated in Fig.~\ref{fig:rapta_pipeline}, RAPTA is a training-time, image-grounded prompt diversification scheme. Given a training pair $(I,p)$ with image $I\in{R}^{H\times W\times 3}$ and base prompt $p$, RAPTA replaces the single caption $p$ by a small set of region-aware textual variants while preserving semantic alignment.
We apply an off-the-shelf detector (Faster R-CNN) to $I$ to obtain candidate boxes, class labels and confidence scores
\[
\{(b_i,c_i,S_i)\}_{i=1}^{N}=\mathrm{Detector}(I),
\]
where $b_i=(x_{i1},y_{i1},x_{i2},y_{i2})$ and $c_i$ is the predicted object class. We merge overlapping boxes with non-maximum suppression (IoU threshold $\tau_{\mathrm{nms}}$), discard low-confidence detections with $S_i\le \tau_b$, and keep the top $M$ proposals. If the detector provides multi-scale features, we run it once; otherwise, optional multi-scale testing is merged by the same NMS.

For each retained box we compute the box center, normalize by the image size, and discretize to a $3{\times}3$ grid $\mathcal{G}$ that yields a coarse position token:
\[
\mathrm{pos}_i=\mathrm{grid}\!\left(\frac{x_{i1}+x_{i2}}{2W},\,\frac{y_{i1}+y_{i2}}{2H}\right)\in\mathcal{G},
\]
with $\mathcal{G}=\{\text{top-left},\ldots,\text{bottom-right}\}$.

Let $\{T_j\}_{j=1}^{J}$ denote a small set of fill-in templates that take $(p,c,\mathrm{pos})$ as inputs. We form a pool of region-aware variants
\[
V=\{p\}\;\cup\;\bigl\{\,T_j\bigl(p,c_i,\mathrm{pos}_i\bigr)\; \big|\; i=1,\ldots,M,\; j=1,\ldots,J \bigr\}.
\]
Two concrete examples are
\emph{``$p$, with a $\langle$c$\rangle$ in the $\langle\mathrm{pos}\rangle$''} and
\emph{``$p$, featuring $\langle$c$\rangle$ and $\langle$c'$\rangle$''} (for a randomly sampled pair of boxes $i\neq i'$).
We keep $J$ small to avoid combinatorial growth; if no reliable box is found, $V=\{p\}$.

To preserve prompt–image coherence, we score each variant $v\in V$ with CLIP:
\[
f_I=\mathrm{CLIP}_{\mathrm{img}}(I),\quad f_v=\mathrm{CLIP}_{\mathrm{text}}(v),\quad
S_v=\frac{f_I^\top f_v}{\|f_I\|\,\|f_v\|}.
\]
Scores are converted to non-negative weights with a temperature $\gamma>0$ and normalized to a sampling distribution:
\[
w_v=\bigl(S_v\bigr)_{+}^{\gamma},\qquad
\pi(v)=\frac{w_v}{\sum_{u\in V} w_u},
\]
where $(\cdot)_{+}=\max(\cdot,0)$. At each iteration we draw one variant $\tilde p\sim\pi(\cdot)$ and encode $e=\mathrm{CLIP}_{\mathrm{text}}(\tilde p)$ to condition the denoiser.

With latent $x_t$ at timestep $t$ and noise $\epsilon\sim\mathcal{N}(0,I)$, the loss remains the standard diffusion objective
\[
\mathcal{L}_{\mathrm{diff}}
={E}_{I,\epsilon,t,\tilde p}\Bigl[\bigl\|\epsilon-\epsilon_{\theta}(x_t,t,e)\bigr\|_2^{2}\Bigr].
\]

Inference-time prompt perturbations show the model only one textual view per image and can drift semantically; in contrast, RAPTA exposes each training image to multiple, image-grounded descriptions across iterations while keeping the conditioning pathway unchanged, reducing reliance on any single caption–image pairing and thereby mitigating memorization.

\begin{algorithm}[t]
\caption{RAPTA: Region-Aware Prompt Augmentation}
\label{alg:rapta}
\begin{algorithmic}[1]
\Require Image $I$, base prompt $p$; detector $\mathrm{Detector}(\cdot)$; template set ${T_j}{j=1}^{J}$; grid $\mathcal{G}$; CLIP encoders; temperature $\gamma$; confidence threshold $\tau_b$; top-$M$ boxes.
\Ensure Sampled prompt $\tilde p$ and embedding $e$.
\State ${(b_i,c_i,S_i)}{i=1}^{N} \leftarrow \mathrm{Detector}(I)$
\State Keep top-$M$ with $S_i>\tau_b$; compute $\mathrm{pos}i\in\mathcal{G}$ from each $b_i$
\State $V \leftarrow {p}$
\For{each kept region $(b_i,c_i)$}
\For{$j\leftarrow1..J$}
\State $V\leftarrow V\cup{T_j(p,c_i,\mathrm{pos}i)}$
\EndFor
\EndFor
\State $f_I\leftarrow \mathrm{CLIP}{img}(I)$
\For{each $v\in V$}
\State $S_v \leftarrow \cos\big( f_I,\mathrm{CLIP}{text}(v)\big)$; $w_v\leftarrow S_v^{\gamma}$
\EndFor
\State $\pi(v)\leftarrow w_v/\sum_{u\in V}w_u$; sample $\tilde p\sim \pi(\cdot)$
\State $e\leftarrow \mathrm{CLIP}_{text}(\tilde p)$; \Return $\tilde p, e$
\end{algorithmic}
\end{algorithm}

\begin{figure*}[t]
\centering
\includegraphics[width=1.04\linewidth]{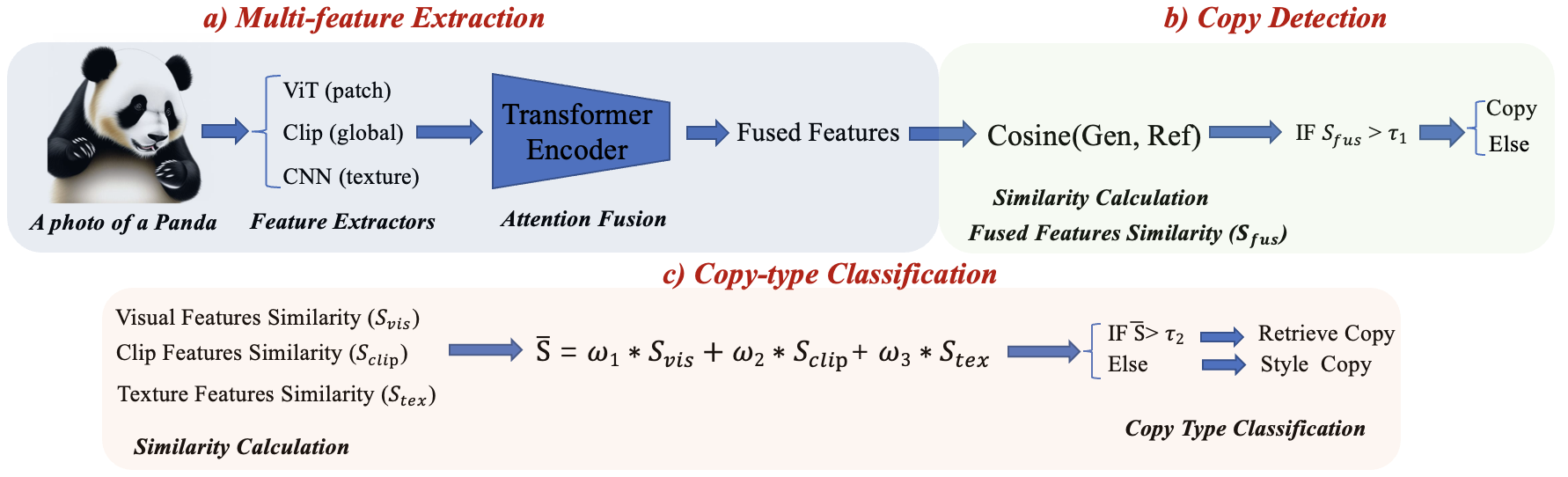}
\caption{\textbf{ADMCD pipeline.}
\textbf{(a)} For an image $X\in\{G,R\}$, three complementary embeddings are extracted: a patch-level visual descriptor $\mathbf f^{\mathrm{vis}}(X)$, a global CLIP descriptor $\mathbf f^{\mathrm{clip}}(X)$, and a texture descriptor $\mathbf f^{\mathrm{tex}}(X)$. After linear projection, the streams are fused by a lightweight Transformer and $\ell_2$-normalized to yield $\hat{\mathbf f}_{\mathrm{fus}}(X)$.
\textbf{(b)} Copy decision uses fused cosine similarity
$S_{\mathrm{fus}}$; we flag \emph{Copy} if $S_{\mathrm{fus}}>\tau_1$.
\textbf{(c)} For flagged pairs we compute stream-wise similarities
$S_{\mathrm{vis}}, S_{\mathrm{clip}}, S_{\mathrm{tex}}$ and the similarity-weighted score
$\bar S$;
\emph{Retrieve/Exact} if $\bar S>\tau_2$, otherwise \emph{Style}. Thresholds $(\tau_1,\tau_2)$ and weights $(\omega_1,\omega_2,\omega_3)$ are fixed from validation sweeps.}
\label{fig:admcd}
\end{figure*}

\subsection{Attention-Driven Multimodal Copy Detection (ADMCD)}
\label{sec:admcd}
As illustrated in Fig.~\ref{fig:admcd}, ADMCD has three parts. (i) A shared, lightweight fusion module maps any input image $I_{G/R}$ (where $G$ is the generated/query image and $R$ is the reference) to a single fused embedding. (ii) A copy decision threshold on the fused cosine similarity detects whether $G$ copies from $R$. (iii) For flagged pairs, a copy–type threshold on a weighted stream score separates retrieve/exact from style copies.

\noindent\textbf{Feature extraction and attention fusion.}
For each image \(X\), we extract three complementary embeddings: a patch-level visual descriptor \(\mathbf f^{\mathrm{vis}}(X)\in{R}^d\), a global CLIP descriptor \(\mathbf f^{\mathrm{clip}}(X)\in{R}^d\), and a texture descriptor \(\mathbf f^{\mathrm{tex}}(X)\in{R}^d\) (see Fig.~\ref{fig:admcd}a).
We apply linear projection to map features onto a common dimensionality-space. We then stack the three feature components and pass them through a small Transformer encoder to perform attention-based fusion. The output is an \(\ell_2\)-normalized fused vector.
\begin{equation}
\begin{aligned}
\hat{\mathbf f}_{\mathrm{fus}}(X)=\mathrm{Attn}\!\left([\mathbf f^{\mathrm{vis}}(X);\mathbf f^{\mathrm{clip}}(X);\mathbf f^{\mathrm{tex}}(X)]\right)\\
\!/\!\bigl\|\mathrm{Attn}(\cdot)\bigr\|_2 \in {R}^d .
\end{aligned}
\end{equation}
% \paragraph*{Copy decision by fused similarity (Fig.~\ref{fig:admcd}b).}
\noindent\textbf{Copy decision by fused similarity.}
As visualized in Fig.~\ref{fig:admcd}b, given \(\hat{\mathbf f}_{\mathrm{fus}}(G)\) and \(\hat{\mathbf f}_{\mathrm{fus}}(R)\), we compute fused-feature cosine similarity
\[
S_{\mathrm{fus}}=\cos\!\big(\hat{\mathbf f}_{\mathrm{fus}}(G),\,\hat{\mathbf f}_{\mathrm{fus}}(R)\big).
\]
A pair is flagged as \emph{Copy} when
\[
S_{\mathrm{fus}}>\tau_1 \quad(\tau_1=0.938\text{ in our setup}),
\]
where \(\tau_1\) is selected on a held-out split to optimize the F1 trade-off.

% \paragraph*{Copy-type classification (Fig.~\ref{fig:admcd}c).}
\noindent\textbf{Copy-type classification  (Fig.~\ref{fig:admcd}c).}
For flagged pairs, we additionally compute stream-wise similarities
\begin{equation}
\begin{aligned}
S_{\mathrm{vis}}=\cos\!\big(\mathbf f^{\mathrm{vis}}(G),\mathbf f^{\mathrm{vis}}(R)\big),\\
\quad
S_{\mathrm{clip}}=\cos\!\big(\mathbf f^{\mathrm{clip}}(G),\mathbf f^{\mathrm{clip}}(R)\big),\\
\quad
S_{\mathrm{tex}}=\cos\!\big(\mathbf f^{\mathrm{tex}}(G),\mathbf f^{\mathrm{tex}}(R)\big),
\end{aligned}
\end{equation}
and form a weighted score consistent with Fig.~\ref{fig:admcd}:
\begin{equation}
\begin{aligned}
\bar S \;=\; \omega_{1}\,S_{\mathrm{vis}} \;+\; \omega_{2}\,S_{\mathrm{clip}} \;+\; \omega_{3}\,S_{\mathrm{tex}}\\
\qquad\bigl(\omega_{1},\omega_{2},\omega_{3} = 0.24,\,0.38,\,0.38\bigr).
\end{aligned}
\end{equation}
By exploiting a similarity threshold, we determine the copy type i.e.,
\begin{equation}
\begin{aligned}
\bar S>\tau_2 \Rightarrow \text{Retrieve Copy},\\
\qquad
\bar S\le\tau_2 \Rightarrow \text{Style Copy},\\
\quad (\tau_2=0.970).
\end{aligned}
\end{equation}

\begin{algorithm}[t]
\caption{ADMCD: Attention-Driven Multimodal Copy Detection (thresholded)}
\label{alg:admcd}
\begin{algorithmic}[1]
\Require Generated image $G$, reference image $R$;
feature extractors (ViT/CLIP/CNN); Transformer encoder;
thresholds $\tau_1,\tau_2$; weights $\omega_1,\omega_2,\omega_3$.
\Ensure Copy flag and type.
\State Extract $\mathbf f^{\mathrm{vis}}(X)$, $\mathbf f^{\mathrm{clip}}(X)$, $\mathbf f^{\mathrm{tex}}(X)$ for $X\!\in\!\{G,R\}$
\State $\hat{\mathbf f}_{\mathrm{fus}}(X)\leftarrow
      \mathrm{Attn}\big([\mathbf f^{\mathrm{vis}}(X);\mathbf f^{\mathrm{clip}}(X);\mathbf f^{\mathrm{tex}}(X)]\big)$;
      $\hat{\mathbf f}_{\mathrm{fus}}(X)\leftarrow
      \hat{\mathbf f}_{\mathrm{fus}}(X)/\|\hat{\mathbf f}_{\mathrm{fus}}(X)\|_2$
\State $S_{\mathrm{fus}}\leftarrow
       \cos\!\big(\hat{\mathbf f}_{\mathrm{fus}}(G),\hat{\mathbf f}_{\mathrm{fus}}(R)\big)$
\If{$S_{\mathrm{fus}}\le \tau_1$}
  \State \textbf{return} Not Copy
\Else
  \State $S_{\mathrm{vis}}\!\leftarrow\!\cos\!\big(\mathbf f^{\mathrm{vis}}(G),\mathbf f^{\mathrm{vis}}(R)\big)$
  \State $S_{\mathrm{clip}}\!\leftarrow\!\cos\!\big(\mathbf f^{\mathrm{clip}}(G),\mathbf f^{\mathrm{clip}}(R)\big)$
  \State $S_{\mathrm{tex}}\!\leftarrow\!\cos\!\big(\mathbf f^{\mathrm{tex}}(G),\mathbf f^{\mathrm{tex}}(R)\big)$
  \State $\bar S \leftarrow \omega_{1} S_{\mathrm{vis}} + \omega_{2} S_{\mathrm{clip}} + \omega_{3} S_{\mathrm{tex}}$
  \If{$\bar S > \tau_2$}
    \State \textbf{return} Copy, type $=$ Retrieve
  \Else
    \State \textbf{return} Copy, type $=$ Style
  \EndIf
\EndIf
\end{algorithmic}
\end{algorithm}

\noindent
Both thresholds \(\tau_1,\tau_2\) and weights \(\omega_{1},\omega_{2},\omega_{3}\) are fixed from validation sweeps and remain constant at test time. The design avoids training any downstream classifier, making ADMCD deployment-ready while retaining strong detection and categorization performance.

\section{Experiments and Results}

\subsection{Experimental setup}
\label{sec:exp_setup}

We curated an evaluation set of 1{,}200 query–reference pairs. It comprises $\sim$25 retrieve/exact pairs, $\sim$200 style-copy pairs, and $\sim$1{,}000 non-copy pairs. We use the following criteria: \emph{retrieve/exact} denotes near-duplicate content and layout (minor photometric changes allowed); \emph{style} preserves global appearance/texture or overall scene style while differing in object identity or geometry; \emph{non-copy} shows no such link to a training item. 

Pairs are constructed by matching each generated/query image $G$ to a training-set reference $R$: for copy cases, $R$ is the mined training image that best matches $G$ (or an exemplar collected from prior studies); for non-copy cases, $R$ is randomly sampled to be class-consistent but non-matching. The distribution is intentionally skewed: truly near-duplicate copies are rare and difficult to elicit from modern models, hence only $\sim$25 high-confidence retrieve/exact examples.

\begin{table*}[t]
\centering
\setlength{\tabcolsep}{4pt}
\renewcommand{\arraystretch}{1.05}
\caption{Result comparison on one query: a generated image $G$ and the Top-5 retrieved references $R_{1..5}$ under different methods.}
\label{tab:qual_top5}
\begin{tabular}{C{3.2cm} C{2.6cm} C{2.6cm} C{2.6cm} C{2.6cm} C{2.6cm}}
\toprule
\multicolumn{1}{c}{} & \multicolumn{5}{c}{Top-5 retrieved references (R)} \\
\cmidrule(lr){2-6}
Generated (G) & $R_1$ & $R_2$ & $R_3$ & $R_4$ & $R_5$ \\
\cmidrule(lr){2-6}
\makecell{\includegraphics[height=20mm,keepaspectratio]{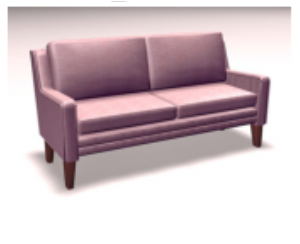}} &
\includegraphics[height=20mm,keepaspectratio]{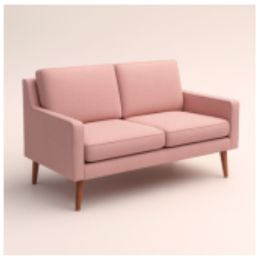} &
\includegraphics[height=20mm,keepaspectratio]{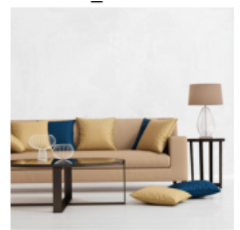} &
\includegraphics[height=20mm,keepaspectratio]{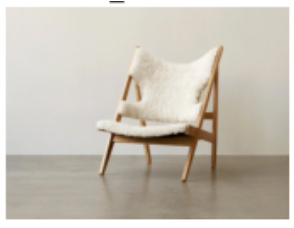} &
\includegraphics[height=20mm,keepaspectratio]{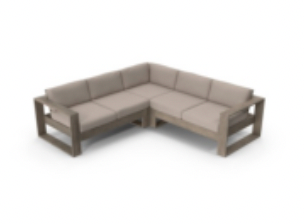} &
\includegraphics[height=20mm,keepaspectratio]{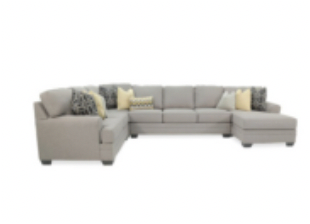} \\
\addlinespace[4pt]
LPIPS~\cite{zhang2018lpips} & 0.349 & 0.503 & 0.447 & \textbf{0.342} & 0.355 \\
ORB~\cite{rublee2011orb}   & 0.174 & 0.136 & 0.156 & \textbf{0.242} & 0.135 \\
SSIM~\cite{wang2004ssim}  & 0.486 & 0.510 & 0.559 & 0.611 & \textbf{0.631} \\
SSCD~\cite{pizzi2022sscd}  & \textbf{0.386} & 0.342 & 0.352 & 0.247 & 0.239 \\
DreamSim~\cite{fu2023dreamsim}  & \textbf{0.755} & 0.646 & 0.732 & 0.681 & 0.697 \\
ADMCD (Ours)  & \textbf{0.959} & 0.859 & 0.878 & 0.868 & 0.850 \\
\bottomrule
\end{tabular}
\end{table*}

We benchmark three conditional diffusion backbones: (i)DCR \cite{somepalli2022forgery,somepalli2023memorize}; (ii) LDM–T2I (CompVis/ldm-text2im-large-256) \cite{compviS_ldm_t2i_256}; (iii) SD2.1-base (stabilityai/stable-diffusion-2-1-base) \cite{sd21_base_hf}.
Regarding the relationship between image similarity and copy detection, we also compare ADMCD to three widely used metrics: LPIPS \cite{zhang2018lpips}, ORB \cite{rublee2011orb}, and SSIM \cite{wang2004ssim}. Each baseline retrieves top-$K$ references for a query $G$ and is thresholded for a binary decision.

\subsection{Threshold and Weight Selection}

All experiments in this section use our in‐house test set comprising 200 copied image pairs and 1000 non‐copied pairs.  Despite its modest size, the dataset exhibits a clear separation between copied and non‐copied examples under our fused similarity metric, allowing us to identify robust thresholds and feature weights.

\noindent\textbf{Copy decision threshold.}
We first sweep the decision threshold \(\tau\) on the cosine similarity \(S_{\mathrm{fus}}\) of the fused features:
\[
\mathrm{is\_copied} = [\,S_{\mathrm{fus}} > \tau\,].
\]
As detailed in the supplementary material (§\ref{sec:TH}), the accuracy–threshold sweep peaks at $\tau=0.938$, which we adopt as the primary copy-detection threshold.

\noindent\textbf{Feature‐weight Heatmap.}
Next, we perform a grid search over weights \((w_{\mathrm{vis}},w_{\mathrm{clip}},w_{\mathrm{tex}})\) (with \(w_{\mathrm{tex}}=1-w_{\mathrm{vis}}-w_{\mathrm{clip}}\)) for the weighted sum
\[
S_{\mathrm{w}} \;=\; w_{\mathrm{vis}}\,S_{\mathrm{vis}}
            + w_{\mathrm{clip}}\,S_{\mathrm{clip}}
            + w_{\mathrm{tex}}\,S_{\mathrm{tex}}.
\]
The weight search heatmap in §\ref{sec:TH} further indicates that $(w_{\mathrm{vis}},w_{\mathrm{clip}},w_{\mathrm{tex}})=(0.24,\,0.38,\,0.38)$ maximizes copy-detection accuracy under the fused-feature threshold.

\noindent\textbf{Copy–type Threshold.}
With the above weights fixed, we re‐evaluate the threshold on the weighted similarity \(S_{\mathrm{w}}\).  Guided by expert human judgments on retrieve‐copy versus style‐copy pairs (multiple annotators, \(n=5\)), we find that a threshold of \(\tau_{\mathrm{w}}=0.970\) cleanly separates retrieve‐copy examples (all \(>0.970\)) from style‐copy ones (all \(<0.970\)).  We adopt \(\tau_{\mathrm{w}}=0.970\) as our final decision boundary for strict copy detection.

\begin{table}[t]
\centering
\caption{Detection vs. conditional diffusion backbones. Copy Rate is computed by ADMCD at threshold $\tau_1$; generation quality is reported on the same sample set.}
\label{tab:main_detection_backbones}
\begin{tabular}{lcccc}
\hline
Method & ADMCD (Ours) & FID & CLIP Score & KID \\
\hline
DCR~\cite{somepalli2022forgery} & 3.2 & 7.9 & 30.5 & 2.9 \\
LDM–T2I~\cite{compviS_ldm_t2i_256} & 5.3 & 10.4 & 33.2 & 3.1 \\
SD2.1-base~\cite{sd21_base_hf} & 7.4 & 8.3 & 27.8 & 3.3 \\
RAPTA(Ours)       & 2.6 & 8.1 & 23.1 & 1.6 \\
\hline
\end{tabular}
\end{table}

\subsection{ Top-5 Retrieval: Cross-metric Comparison}
\label{sec:qual_top5_example}

Table~\ref{tab:qual_top5} compares five commonly used detectors on the \emph{same} candidate set: 
three traditional metrics that many recent works adopt as copy detectors (LPIPS distance, ORB keypoint matching, and SSIM), 
and two recent strong baselines, SSCD (a learned global fingerprint for copy detection) and DreamSim (a perceptual similarity model). 
For each retrieved candidate $R_k$ we report the score between $G$ and $R_k$ (LPIPS is a distance, others are similarities).
Under identical retrieval, SSCD and DreamSim often identify a very plausible nearest neighbor; however, the top-$k$ scores for these single-encoder methods tend to be close to each other, making thresholding less decisive. The  traditional metrics also emphasize different single views: LPIPS is biased toward texture/color similarity, ORB favors repeatable keypoints and rigid geometry, and SSIM is dominated by luminance/contrast and local structure—hence each selects a different “best” reference.

Our fused similarity $S_{\mathrm{fus}}$ yields a clearer and more stable ranking: it concentrates probability mass on the visually most faithful neighbor while maintaining consistent scores for the next few candidates.  This behavior stems from how ADMCD constructs the score: we \emph{attentively fuse} three complementary cues—ViT patch-level geometry, CLIP global semantics, and CNN texture—so when one cue is ambiguous (e.g., texture-only matches for LPIPS, keypoint sparsity for ORB, or illumination shifts for SSIM), the other cues compensate. In contrast, DreamSim optimizes general perceptual agreement rather than data-replication specifically, and SSCD’s single-stream global embedding can miss fine-grained layout or style differences and yields smaller margins among near-ties. Moreover, none of the baselines separate retrieve/exact vs.\ style copying, whereas ADMCD uses a second threshold to make this distinction, leading to a more actionable decision rule in downstream analysis.

\subsection{Main results: copy mitigation with RAPTA and detection with ADMCD}
\label{sec:results}

As summarized in Table~\ref{tab:main_detection_backbones}, our combined method—RAPTA for training-time mitigation and ADMCD for evaluation—achieves the lowest copy rate (2.6), reducing it by 0.6, 2.7, and 4.8 absolute points relative to DCR (3.2), LDM–T2I (5.3), and SD2.1-base (7.4), i.e., 18.8\%, 50.9\%, and 64.9\% relative reductions. Copy rate is measured with ADMCD under a fixed threshold protocol and a shared reference gallery. RAPTA lowers memorization while preserving visual quality: FID remains comparable to the strongest baselines (8.1 versus 7.9 and 8.3, and better than 10.4), and KID improves (1.6 versus 2.9–3.3). CLIP Score is lower (23.1 versus 27.8–33.2), reflecting a trade-off in text–image similarity when suppressing replication, yet overall perceptual quality is maintained alongside a marked drop in copying.

\subsection{Robustness to common attacks}

\begin{table}[t]
\centering
\caption{Similarity between a generated image $G$ and a fixed reference $R$ under noise perturbations. Column Original reports the score for $(G,R)$ without any perturbation; the other columns report scores after perturbing \emph{only $G$} with Gaussian noise, Gaussian blur, Poisson noise, salt--pepper, and speckle. Higher is better except LPIPS (distance).}
\label{tab:robust_photometric}
\setlength{\tabcolsep}{2.5pt}
\renewcommand{\arraystretch}{1.05}
\begin{tabular}{lcccccc}
\hline
Method & Original & Gauss N. & Gauss B. & Poisson & S\&P & Speckle \\
\hline
LPIPS~\cite{zhang2018lpips} & 0.233 & 0.444 & 0.335 & 0.375 & 0.612 & 0.569 \\
ORB~\cite{rublee2011orb}   & 0.276 & 0.242 & 0.155 & 0.282 & 0.172 & 0.184 \\
SSIM~\cite{wang2004ssim}  & 0.677 & 0.504 & 0.664 & 0.591 & 0.389 & 0.407 \\
SSCD~\cite{pizzi2022sscd}  & 0.680 & 0.594 & 0.443 & 0.429 & 0.485 & 0.407 \\
DreamSim~\cite{fu2023dreamsim}  & 0.857 & 0.781 & 0.714 & 0.691 & 0.689 & 0.707 \\
ADMCD(Ours)  & 0.974 & 0.923 & 0.940 & 0.929 & 0.871 & 0.894 \\
\hline
\end{tabular}
\end{table}

\begin{table}[t]
\centering
\caption{Similarity between a generated image $G$ and a fixed reference $R$ under geometric perturbations applied only to $G$: 20\% crop, horizontal/vertical flips, 10\% occlusion, and $30^\circ$ rotation. Higher is better except LPIPS (distance).  $G$ and  $R$ are the same as in Table~\ref{tab:robust_photometric}.}
\label{tab:robust_geometric}
\setlength{\tabcolsep}{2.5pt}
\renewcommand{\arraystretch}{1.05}
\begin{tabular}{lccccc}
\hline
Method & Crop20\% & Flip H & Flip V & Occlude10\% & Rotate$30^\circ$ \\
\hline
LPIPS~\cite{zhang2018lpips} & 0.335 & 0.373 & 0.519 & 0.315 & 0.615 \\
ORB~\cite{rublee2011orb}   & 0.232 & 0.221 & 0.243 & 0.110 & 0.276 \\
SSIM~\cite{wang2004ssim}  & 0.570 & 0.556 & 0.427 & 0.642 & 0.207 \\
SSCD~\cite{pizzi2022sscd}  & 0.577 & 0.404 & 0.464 & 0.391 & 0.489  \\
DreamSim~\cite{fu2023dreamsim}  & 0.617 & 0.524 & 0.564 & 0.691 & 0.689 \\
ADMCD(Ours)  & 0.970 & 0.886 & 0.857 & 0.748 & 0.939 \\
\hline
\end{tabular}
\end{table}

As summarized in Table~\ref{tab:robust_photometric} and Table~\ref{tab:robust_geometric}, our fused similarity remains high and stable across ten perturbations. Under noise-based corruptions, ADMCD stays in the range 0.871--0.974, with only mild drops from 0.974 to 0.871 (S\&P) and 0.894 (SP). DreamSim is the next best, but decreases under blur and noise. SSCD's global fingerprint is relatively stable on mild noise yet degrades on blur and speckle. Classical metrics behave as expected: LPIPS (distance) rises strongly with impulsive noise, ORB collapses on blur and speckle, and SSIM drops on S\&P.  For geometric robustness, when only $G$ is perturbed by crop, flips, occlusion, and $30^\circ$ rotation, ADMCD still preserves the match (0.748--0.970). SSCD suffers larger drops under flips and occlusion. Baselines are less stable overall: LPIPS distance increases on rotation, ORB falls to 0.110 under occlusion, and SSIM drops to 0.207 on rotation.

These trends underscore why ADMCD is robust: its \emph{three-stream} design combines patch-level geometry (ViT) for spatial anchors, global semantics (CLIP) for invariance to color/illumination, and texture cues (CNN) for resilience to noise/blur. Attention-based fusion down-weights a failing stream so no single weakness dominates the score. Consequently, ADMCD aligns better with human-perceived similarity and maintains a single pair of thresholds $(\tau_1,\tau_2)$ across both photometric and geometric conditions.

\section{Conclusion}
In this work we have presented a new framework for both mitigating and detecting unintended memorization in text-to-image diffusion models.  On the generation side, we introduced Region-Aware Prompt Augmentation (RAPTA), which leverages a pretrained object detector to produce semantically grounded prompt variants and randomly samples one per training iteration.  RAPTA preserves the core semantics of each prompt while injecting controlled diversity, reducing overfitting without degrading visual fidelity.  On the detection side, we proposed Attention-Driven Multimodal Copy Detection (ADMCD), which fuses complementary patch-level, global semantic and texture features via a lightweight transformer and applies a simple threshold rule to flag copied content—entirely without requiring large labeled corpora.  

Extensive experiments on large-scale image collections demonstrate that RAPTA consistently enhances prompt–image consistency and model robustness under diverse training prompts, and that ADMCD outperforms single-view similarity metrics in a zero-training regime, achieving reliable copy-detection in practice.  Our two modules integrate seamlessly into existing diffusion pipelines, offering a balanced solution to the twin challenges of generative quality and intellectual property safety.

\bibliographystyle{IEEEtran}  
\bibliography{ref}            

\appendices

\section{Supplementary Material}

\subsection{Threshold and Weight Selection}
\label{sec:TH}
\begin{figure}[ht]
  \centering
  \includegraphics[width=0.8\linewidth]{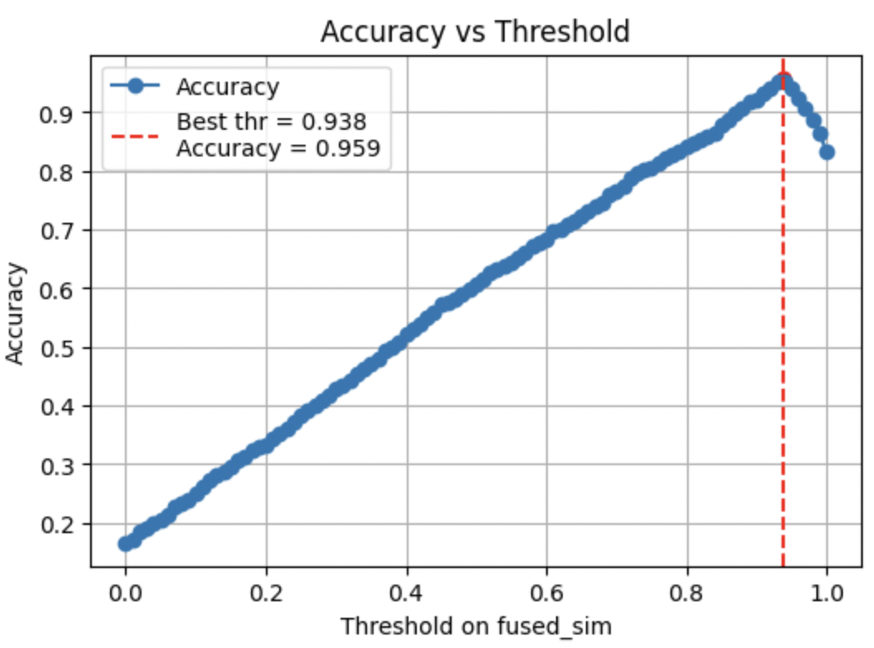}
  \caption{Accuracy vs.\ threshold on fused similarity \(S_{\mathrm{fus}}\).  Best accuracy is achieved at \(\tau=0.938\).}
  \label{fig:threshold_fused}
\end{figure}

Figure~\ref{fig:threshold_fused} plots accuracy versus \(\tau\).  The peak occurs at \(\tau=0.938\), which we thus adopt as our primary copy‐detection threshold.

\begin{figure}[ht]
  \centering
  \includegraphics[width=0.8\linewidth]{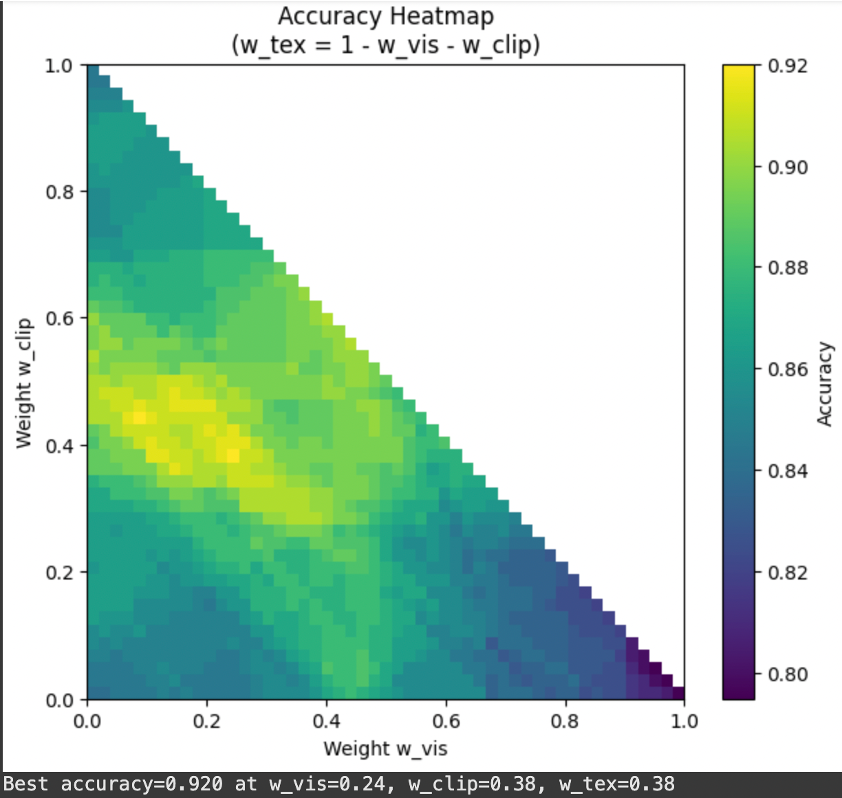}
  \caption{Accuracy heatmap over \((w_{\mathrm{vis}},w_{\mathrm{clip}})\) with \(w_{\mathrm{tex}}=1-w_{\mathrm{vis}}-w_{\mathrm{clip}}\).  Peak at \((0.24,0.38,0.38)\).}
  \label{fig:weight_heatmap}
\end{figure}

Figure~\ref{fig:weight_heatmap} shows the resulting accuracy heatmap.  We select \((w_{\mathrm{vis}},w_{\mathrm{clip}},w_{\mathrm{tex}})=(0.24,\,0.38,\,0.38)\), which maximizes copy‐detection accuracy under the fused‐feature threshold.

\begin{figure*}[t]
\centering
\includegraphics[width=0.8\linewidth]{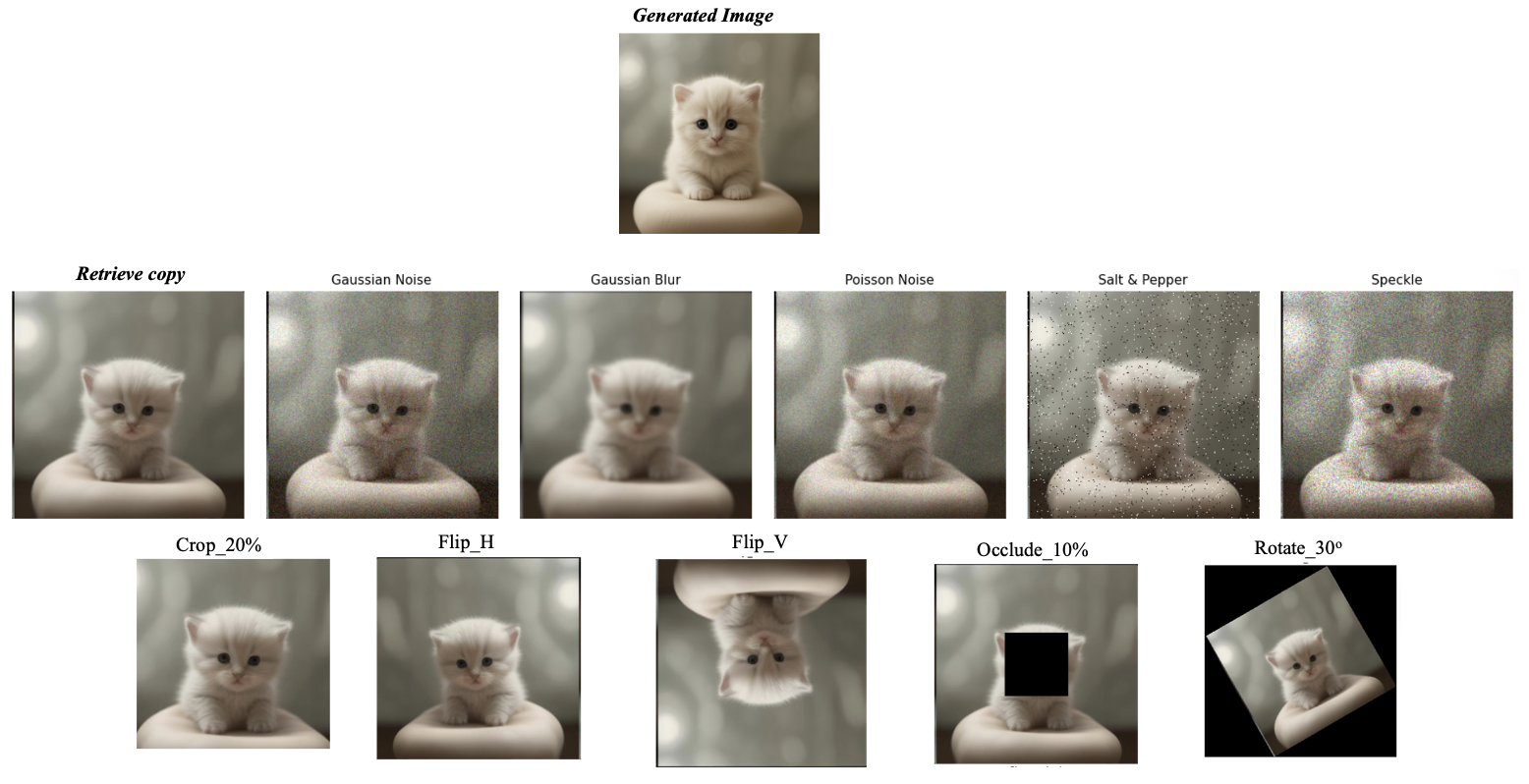}
\caption{\textbf{Robustness examples.} Top: generated query $G$. Bottom: the clean retrieved copy $R$ and ten corrupted variants—Gaussian noise, Gaussian blur, Poisson noise, salt--pepper, speckle, 20\% crop, horizontal/vertical flips, 10\% occlusion, and $30^\circ$ rotation. Our fused-feature similarity $S_{\mathrm{fus}}$ remains high and continues to select the correct match across both photometric and geometric perturbations. Numerical scores for each method appear in Tables~\ref{tab:robust_photometric} and \ref{tab:robust_geometric}.}
\label{fig:robust_examples}
\end{figure*}

\subsection{Qualitative robustness under common attacks}
\label{sec:robust_examples}
Figure~\ref{fig:robust_examples} illustrates how our detector behaves when the matched reference is corrupted by typical photometric and geometric perturbations. The top shows the generated query $G$; the bottom two rows show the clean retrieved copy $R$ and ten corrupted variants: Gaussian noise, Gaussian blur, Poisson noise, salt--pepper noise, speckle noise, 20\% center crop, horizontal flip, vertical flip, 10\% occlusion, and a $30^\circ$ rotation. Across these manipulations the fused-feature similarity $S_{\mathrm{fus}}$ remains high and the correct match is consistently preserved, indicating that the attention-fused embedding is largely invariant to moderate noise, blur, and monotonic intensity shifts, and is resilient to small viewpoint/geometry changes. Quantitatively, Tables~\ref{tab:robust_photometric} and \ref{tab:robust_geometric} show that our score stays within a tight band under photometric noise and blur and degrades gracefully under geometry (e.g., still reliable at 10\% occlusion and $30^\circ$ rotation), whereas LPIPS, ORB, and SSIM vary substantially with texture, keypoints, and luminance biases. These examples support that ADMCD aligns with human-perceived similarity even when images are corrupted by common real-world artifacts.

\subsection{Single-image copy verification and robustness}
\label{sec:perimage_robust}

\begin{figure*}[t]
\centering
\includegraphics[width=0.8\linewidth]{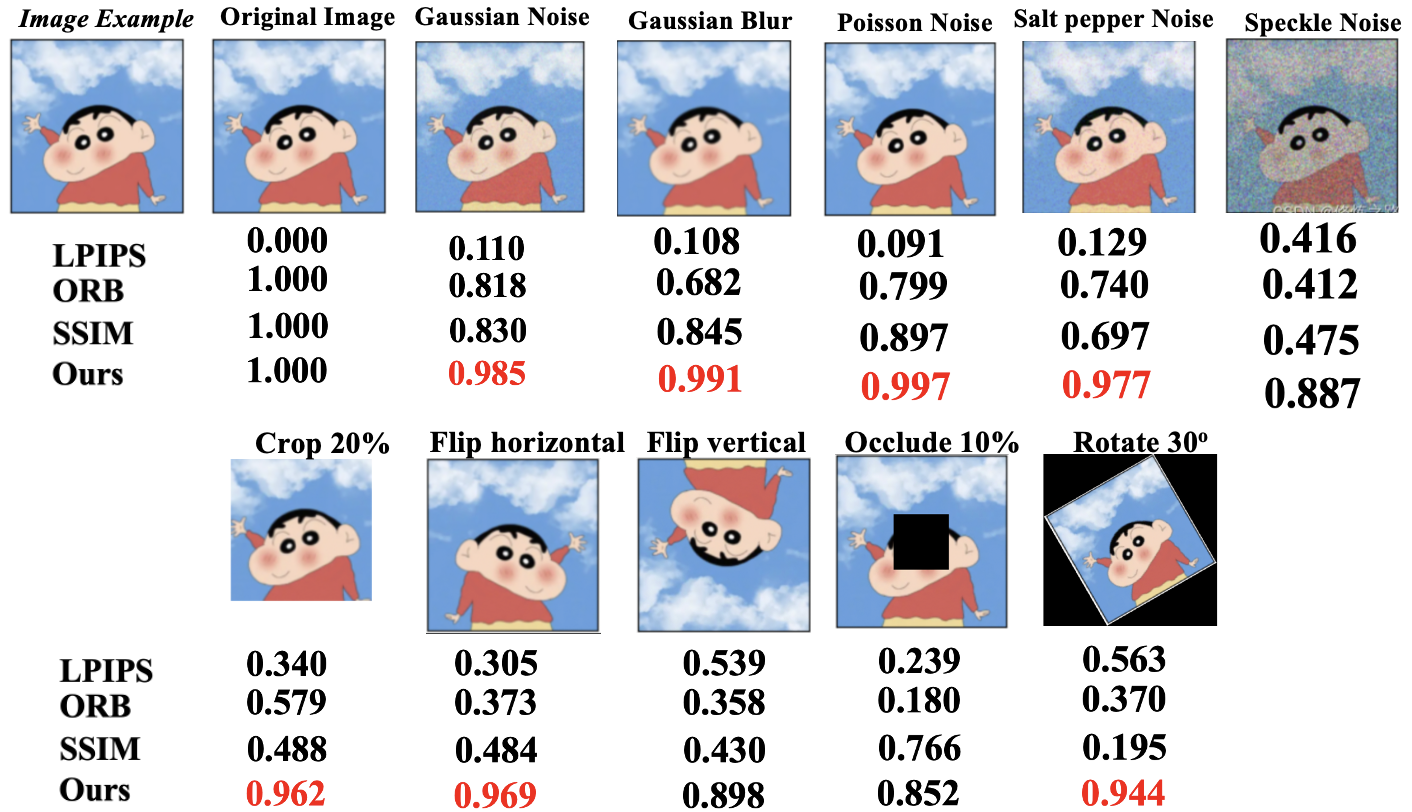}
\caption{Per-image robustness under ten common image-based attacks. For a query $G$ and its retrieved copy $R$, we apply Gaussian noise/blur, Poisson, salt–pepper, speckle, 20\% crop, horizontal/vertical flips, 10\% occlusion, and a $30^\circ$ rotation.}
\label{fig:same}
\end{figure*}

To further validate the accuracy of our copy-rate estimates, we conduct a per-image study on a query $G$ and its retrieved training copy $R$. We first confirm the copy decision on the clean pair ($S_{\mathrm{fus}}=1.000$ for this example), and then corrupt $R$ with ten common attacks: Gaussian noise, Gaussian blur, Poisson noise, salt--pepper noise, speckle noise, 20\% center crop, horizontal flip, vertical flip, 10\% occlusion, and a $30^\circ$ rotation (see Fig.~\ref{fig:same}). Recomputing the similarities shows that our fused score remains high across all perturbations, with values $0.985, 0.991, 0.997, 0.977, 0.887, 0.962, 0.969, 0.898, 0.852,$ and $0.944$ respectively—i.e., always above $0.85$, so the copy decision $S_{\mathrm{fus}}>\tau_1$ is preserved. In contrast, single-view metrics fluctuate strongly: LPIPS rises to $0.416$ (speckle) and $0.563$ (rotation), ORB drops to $0.180$ under occlusion, and SSIM falls to $0.195$ under $30^\circ$ rotation. These observations indicate that our method‘s decision is stable under both photometric and geometric perturbations, supporting that the reported copy rate reflects genuine replication rather than fragile, attack-sensitive matching.

\end{document}